\documentclass[runningheads]{llncs}
\usepackage{graphicx}
\graphicspath{{./assets/}}
\usepackage{float}
\usepackage[T1]{fontenc}
\usepackage{listings}
\usepackage{xcolor}
\usepackage{tabularx}
\usepackage{textcomp}
\usepackage{orcidlink}
\usepackage{soul}
\usepackage{todonotes}
\usepackage{hyperref}
\usepackage{booktabs}
\usepackage[table]{xcolor}

\definecolor{codebg}{RGB}{34,45,62}
\definecolor{codetext}{RGB}{230,230,230}

\newcommand{\kMeansClusters}{2}
\newcommand{\pdfAttributesCount}{35}

\newcommand{\scrappedProperties}{3,965}
\newcommand{\totalPropertiesForMLPipeline}{2,766}
\newcommand{\textOnlyPDFs}{2,781}

\newcommand{\textOnlyPDFsPercentage}{70.1\%}
\newcommand{\excudedPDFsPercentage}{29.9\%}

\newcommand{\cosineEuclidean}{0.82}
\newcommand{\cosineManhattan}{0.63}
\newcommand{\kMeanSilhouette}{0.2088}

\begin{document}

\title{Structured Data Extraction from Real Estate Documents using Clustering, Classification, and Large Language Models}
\author{Muhammad Assad Shehbaz \and
 Carlos Francisco Moreno-García\orcidlink{0000-0001-7218-9023}}
 \institute{Robert Gordon University,  Garthdee Rd, Aberdeen, UK \and
 \email{m.shehbaz,c.moreno-garcia@rgu.ac.uk} \\
 \url{https://www.rgu.ac.uk}}

 \titlerunning{Real Estate Clustering}
 \authorrunning{Shehbaz et al.}

\maketitle

\begin{abstract}
Real estate property listings expose structured metadata through the API. Still, the richest property-level information (i.e., legal status, structural condition, utility supplies, heating systems) sits in attached questionnaire documents that no automated system currently processes at scale. These documents are heterogeneous. Some are digitally generated with selectable text, others are scanned physical forms. There are even more complex layouts that contain checkbox annotations that defeat conventional text extraction. In this paper, we present an end-to-end pipeline for acquiring, classifying, and extracting structured data from selectable text documents. The pipeline was applied to $\scrappedProperties$ questionnaire documents collected from a live property platform via reverse-engineered REST APIs. First, we classified each document into one of three structural categories (\texttt{text\_only}, \texttt{scanned}, and \texttt{special\_char}), then extracted $\pdfAttributesCount$ predefined property attributes from eligible documents using DeepSeek R1 as the Large Language Model, prompted to return a structured JSON object. All $\textOnlyPDFs$ submitted documents were processed successfully, producing a final dataset of $\totalPropertiesForMLPipeline$ unique property records. Downstream validation confirmed the data quality. Cosine similarity matching achieves a Jaccard consistency score of $\cosineEuclidean$, and K-Means clustering produces interpretable market segments with a silhouette score of $\kMeanSilhouette$. Results show that the proposed extraction from each property document is both feasible and reliable at this scale.
\end{abstract}


\section{Introduction}
\label{sec:intro}
When a property is listed for sale in the United Kingdom (UK), the seller is required to complete a detailed questionnaire covering everything from the central heating system to the presence of asbestos, legal disputes affecting the title, and the names of utility suppliers~\cite{lawsociety_ta6}. This information is usually captured in a document in Portable Document Format (PDF). These documents are neither stored in any database nor accessible through any Application Programming Interface (API), nor indexed by any search engine. They exist as unstructured, variably formatted documents and are largely ignored by automated systems.

Most real estate platforms expose basic structured listing data through some kind of proprietary APIs, such as price, number of bedrooms, floor area, and geographic coordinates~\cite{rightmove_rtdf}. That data is clean and easy to work with. The questionnaire PDFs attached to each listing contain a second, richer layer of information that never makes it into any structured pipeline. For instance, a seller may describe central heating as \textit{gas central heating}, \textit{GCH}, \textit{mains gas}, or \textit{combine boiler system}, which can confuse a potential buyer. Traditional parsing approaches fail on free-form variation. Regular expressions, template matching, and rule-based extractors all depend on fixed patterns and predefined vocabularies. A rule written for one form will silently miss the others, and extending the rule set to cover every variant is neither scalable nor reliable against phrasings not seen at design time. No hand-crafted rule set survives contact with real data. However, recent work shows that LLMs can extract structured fields from free-form text without any document-specific rules~\cite{dunn2024}. Given a page of property disclosure text, a well-prompted model can identify the relevant attributes and return them as a valid JSON object. Whether this holds reliably across thousands of real documents from a live platform, and whether the output is useful downstream, is what we set out to test.

In this paper, we present an end-to-end pipeline for acquiring, classifying, and extracting structured information from real estate questionnaire PDFs at scale, applied to $\scrappedProperties$ documents. The pipeline makes the following contributions:

\begin{enumerate}
    \item A data acquisition system built on reverse-engineered REST, yielding $\scrappedProperties$ questionnaire PDFs alongside structured listing metadata.
    
    \item A model that classifies questionnaire documents into three categories \texttt{scanned}, \texttt{text\_only}, and \texttt{special\_char}, routing each to an appropriate processing pathway.
    
    \item An LLM-based extraction pipeline using DeepSeek R1 model~\cite{deepseek2025} via the Groq inference API~\cite{groqapi}, extracting $\pdfAttributesCount$ predefined property attributes per document as structured JSON, with a 100\% success rate across the $\textOnlyPDFs$ eligible documents.
    
    \item A downstream validation study confirming that the extracted features support similarity matching, market segmentation, and multi-criteria ranking, providing empirical evidence of data quality.
\end{enumerate}

The remainder of the paper is organised as follows. Section~\ref{sec:back} reviews related work on PDF extraction, LLMs for document understanding, and real estate document analysis. Section~\ref{sec:data} describes the data acquisition pipeline. Section~\ref{sec:pipeline} presents the document processing pipeline in detail. Section~\ref{sec:valid} reports the validation experiments. Section~\ref{sec:results} discusses the results. Section~\ref{sec:concs} concludes with directions for future work.

\section{Background}
 \label{sec:back}
\subsection{PDF Document Processing and Information Extraction}
 Extracting structured information from PDF documents is a long-standing challenge in document analysis. The difficulty is not any single technical limitation but from the fundamental heterogeneity of the format itself. A PDF may contain digitally typeset text with a selectable layer, a scanned page image with no text layer, or a hybrid of both within the same file. Property questionnaire documents add a further complication, with form fields, checkbox annotations, and non-standard character encodings being common, and each breaks conventional parsing differently.
 
Traditional approaches rely on rule-based text extraction libraries such as pdfplumber~\cite{pdfplumber} and PyMuPDF~\cite{pymupdf}. These work well on clean, digitally generated PDFs. They fail systematically on documents with inconsistent layouts or encoded form fields, a limitation well established in the PDF extraction literature~\cite{bast2017,subramani2020}.
For scanned documents, OCR provides a means to extract text, but introduces its own error profile. Blurred stamps, skewed pages, and handwritten annotations all degrade output quality, and modern deep learning OCR approaches have been shown to substantially outperform classical engines on noisy real-world documents~\cite{subramani2020}. Any pipeline operating over a real-world corpus must therefore classify documents by type before attempting extraction.

 
Layout-aware models such as LayoutLM~\cite{layoutlm2020} and LayoutLMv3~\cite{layoutlmv32022} address some of these limitations by jointly modelling text content and spatial position. LayoutLMv3 achieves strong results on key information extraction benchmarks by treating document layout as a first-class input signal. These models require fine-tuning on labelled datasets, however, which limits their use in domains, like property questionnaires, where no labelled training data exists.
 
\subsection{Large Language Models for Document Understanding}
 
The emergence of instruction-following large language models has changed what is possible in document information extraction. Rather than training a task-specific model on labelled examples, a well-designed prompt can instruct a general-purpose LLM to identify target fields, handle terminological variation, and return output in a specified schema without any document-specific fine-tuning. Dunn et al.~\cite{dunn2024} demonstrated this on scientific corpora, showing that LLMs extract structured records from free-form documents and return them reliably as JSON objects. Results generalise across domains.
 
The central challenge in applying LLMs to batch extraction is output consistency. Left unconstrained, a model may return the same date as \textit{2024-06-03} in one response and \textit{3rd June 2024} in another. Schema enforcement, requiring the model to return a fixed set of named fields as strings, with \texttt{null} for absent values, addresses this directly and is now standard practice in production document extraction pipelines~\cite{hybridocr2025}. At low temperature settings, modern models produce stable outputs across structurally similar documents. Hallucination is mitigated by prompting for \texttt{null} rather than inference, and by validating output before database insertion. Evaluations across domains, including health records~\cite{bmjllm2025}, confirm that well-constrained prompts achieve high consistency, though performance degrades on very short or poorly formatted documents.
 
\subsection{Document Analysis in Real Estate}
 
Automated extraction from real estate documents has received limited attention in the document analysis literature. The ICDAR~2023 competition on structured text extraction from visually-rich documents~\cite{icdar2023} highlighted the broader difficulty. Even on constrained enterprise document benchmarks, end-to-end systems struggle with format variability and zero-shot generalisation. Property questionnaires are a specific instance of this challenge. They are semi-structured forms with consistent field semantics but highly variable surface presentation across sellers, agents, and time periods. No published pipeline addresses this document type directly.
 
The closest related work is Zhao et al.~\cite{zhao2024}, who apply transformer-based LLMs to information extraction from real estate transaction contracts. They found that general-purpose LLMs handle domain-specific legal and financial language without fine-tuning. Terminological variation is identified as the primary failure mode for rule-based approaches, consistent with our observations across the ASPC questionnaire corpus. Their work targets contract documents rather than seller questionnaires, and does not address the document classification problem or extraction at the scale considered here.


\subsection{Gaps and Challenges}

Several gaps remain unaddressed in the existing literature. First, no published work tackles seller questionnaire PDFs as a document type. Prior work on real estate documents focuses on
contracts~\cite{zhao2024} and transaction records, which have a predictable structure and professional authorship. Seller questionnaires differ on both counts. They are completed by
non-expert individuals, producing inconsistent terminology, irregular phrasing, and variable layout across thousands of documents.

Second, document heterogeneity has not been addressed at the pipeline level in this domain. Real-world property questionnaire corpora contain digitally generated text, scanned physical forms, and checkbox-annotated layouts in the same collection. The ICDAR~2023 competition on heterogeneous document understanding~\cite{icdar2023} confirms that mixed-type corpora remain an open challenge: existing systems apply a single extraction strategy without first classifying documents by type, which fails silently on anything other than clean, digitally generated text~\cite{bast2017}.

Third, the extraction scale has not been demonstrated on this document type. Zhao et al.~\cite{zhao2024} operate on small curated contract sets. The ICDAR~2023 benchmarks~\cite{icdar2023} use controlled document collections. Neither addresses batch extraction across thousands of heterogeneous documents from a live platform.

\section{Data Acquisition}
\label{sec:data}
This section describes the data acquisition pipeline, from API access and iterative request handling through to PDF downloading and initial corpus statistics. This is a reproducible pipeline, operating against the Aberdeen Solicitors Property Centre (ASPC) platform\footnote{\url{http://aspc.co.uk}}.

\subsection{Data Source: ASPC Real Estate Platform}
ASPC is a regional property listing platform operated by a consortium of solicitors from the city of Aberdeen, Scotland, UK, providing public access to residential property listings across the area. It was selected as the data source for this study because it serves a geographically coherent market suitable for regional analysis, and its property listings include downloadable questionnaire PDFs containing rich, unstructured property information not exposed through standard listing views.

Unlike major national platforms such as \textit{Rightmove}\footnote{\url{https://www.rightmove.co.uk/}} or \textit{Zoopla}\footnote{\url{https://www.zoopla.co.uk/}}, ASPC does not publish an official public API. Data acquisition, therefore, required reverse-engineering the platform's internal REST API endpoints, described below.

\subsection{REST API Reverse Engineering and Endpoint Discovery}
Chrome DevTools Network Inspector~\cite{chromedevtools} was used to monitor HTTP traffic during manual browsing sessions on the ASPC platform. This analysis revealed a set of undocumented internal REST API endpoints returning structured JSON responses. Three endpoints were identified as relevant and are summarised in Table~\ref{table:table1}.

\begin{table}[h]
\caption{ASPC REST APIs used for data acquisition}
\centering
\begin{tabularx}{0.95\linewidth}{lXl}
\toprule
\textbf{Endpoint} & \textbf{Response} & \textbf{Format} \\
\midrule

\rowcolor{gray!15}
Search & Paginated listing summaries across search results & JSON \\

Detail & Extended property attributes beyond search results & JSON \\

\rowcolor{gray!15}
Document & URLs to downloadable PDF documents per listing & PDF \\

\bottomrule
\end{tabularx}
\label{table:table1}
\end{table}

The Property Search API supported server-side pagination, enabling systematic retrieval of all available listings. For each page, a GET request was issued using Python's \textit{requests} library~\cite{requests}
and the JSON response was parsed to extract per-property metadata. The Property Detail API was subsequently called for each property ID to retrieve supplementary attributes not included in search result summaries. The Property Document API was called to download the property questionnaire PDFs associated with each listing.

\subsection{Request Strategy, Metadata Extraction and PDF Acquisition}
The ASPC platform requires an authenticated session to access the property PDF reports. Therefore, data acquisition proceeded in three steps within a single authenticated session. First, a \texttt{requests.Session()} object was used to submit a POST request to the ASPC login endpoint, returning a persistent authentication cookie (\texttt{aspc.live.auth.cookie}) that authorised all subsequent API calls and PDF downloads. Second, the Property Search API was called iteratively with a page size of 100 listings per request. For each result, the Property Detail API was called individually to retrieve the full attribute set, including price, floor area, bedroom and bathroom counts, and council tax band. Also, $\pdfAttributesCount$ boolean property attributes were inserted directly into a Microsoft SQL Server database. Failed detail requests were skipped via a \texttt{continue} statement without interrupting the batch. Third, each property detail response was queried for property questionnaire PDFs. Documents were downloaded via authenticated streaming GET requests and saved to subdirectories, named by their SQL database identifier to maintain a traceable link between each PDF and its corresponding property record.

\begin{figure}
\centering
\includegraphics[width=\textwidth]{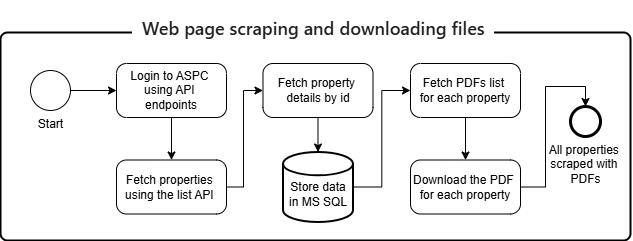}
\caption{Web scraping and PDF acquisition pipeline}
\label{fig:pipeline}
\end{figure}

The complete acquisition process yielded $\scrappedProperties$ unique property listings and associated PDF reports. Some listings included multiple document types (e.g., property questionnaire, energy performance certificate, floor plan), while others had a single questionnaire PDF. Non-questionnaire PDFs were filtered out during the classification stage described in Section~\ref{sec:pipeline}.

Acquisition was conducted for non-commercial academic research purposes only. No personally identifiable information beyond property addresses (publicly listed on the platform) was collected or retained~\cite{khder2021}.

\section{Document Processing Pipeline}
\label{sec:pipeline}
The document processing pipeline comprises two sequential stages. PDF classification and LLM-based structured feature extraction. Designed to transform heterogeneous, unstructured property questionnaire PDFs into a clean, machine-readable feature dataset.

\subsection{Stage 1: PDF Classification}
Before extraction, each questionnaire PDF was classified into one of three structural categories. The classifier first measured the average extractable text length per page using \texttt{pdfplumber}~\cite{pdfplumber}. Documents with an average below 50 characters were classified as \texttt{scanned}, indicating image-rendered pages with no selectable text layer. Next, \texttt{PyMuPDF}~\cite{pymupdf} inspected the text layer for the presence of checkbox or tick marks. As a back-up, pages 2–4 of ambiguous documents were converted to images using \texttt{pdf2image} and scanned with \textit{Tesseract OCR}~\cite{tesseract}, only to detect these same symbols, not to extract content. Documents containing such patterns were classified as \texttt{special\_char}. All remaining text-bearing documents were classified as \texttt{text\_only}.

Of the $\scrappedProperties$ questionnaire PDFs processed, $\textOnlyPDFs$ ($\textOnlyPDFsPercentage$) were classified as \texttt{text\_only} and forwarded to the extraction stage. The remaining document types were excluded from extraction in this study.

\begin{figure}
\centering
\includegraphics[width=\textwidth]{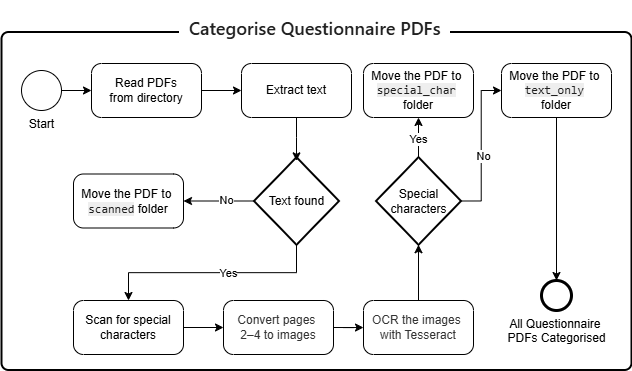}
\caption{Questionnaire PDF classification}
\label{fig:pdf_classification}
\end{figure}


\subsection{Stage 2: LLM-Based Structured Feature Extraction}
All \texttt{text\_only} PDFs were processed using \texttt{pdfplumber} to extract their full raw text. No preprocessing or page selection was applied at this stage. The complete document text was passed directly to the LLM, allowing the model to handle terminological variation and layout inconsistency without rule-based normalisation. This approach was chosen after rule-based parsing was found inadequate due to high variability in how sellers described the same property attributes. For example, central heating appeared variously as \textit{gas central heating}, \textit{GCH}, \textit{mains gas}, and \textit{combine boiler system} across documents.

\subsubsection{Model Selection and Configuration}
The DeepSeek R1 model~\cite{deepseek2025} ($deepseek-r1-distill-llama-70b$), accessed via the Groq inference API, was selected for its strong instruction-following capability and low-latency throughput~\cite{groqapi}. The model was configured with a temperature of $0.3$, top-p of $0.95$, and a maximum of $1,024$ output tokens per request.

\subsubsection{Prompt Design}
A structured prompt instructed the model to extract $\pdfAttributesCount$ predefined property attributes and return them as a single valid JSON object. The prompt enforced three constraints. Return only a valid JSON object with no preamble, represent all values as strings, and include all $\pdfAttributesCount$ fields in every response, using \texttt{null} for absent values. This produced a consistent schema across all outputs. The target attributes span six categories. Legal and ownership status, structural condition, heating and energy, utility supplies, utility suppliers, and shared charges. The low-temperature setting (0.3) reduced output variance among structurally similar documents, improving consistency for features that appeared frequently across the corpus.

\subsubsection{Extraction Pipeline and Error Handling}
For each document, the extracted text was embedded in the prompt and submitted to the Groq API. The JSON object was isolated from the response using a regular expression (re.search(r’\textbackslash \{.*\textbackslash\}’, response, re.DOTALL)) and parsed with \texttt{json.loads()}. The \texttt{property\_id}, matching the SQL database ID used during data acquisition, was appended to each record before insertion into \texttt{property\_questionnaire\_info}. On successful insertion, the source PDF was moved to an \texttt{extracted} folder. On any exception, including JSON parse failures, database errors, or a rollback was triggered, the file was moved to an \texttt{error} folder. This preserved full traceability without interrupting the batch. All $\textOnlyPDFs$ submitted documents were successfully processed and inserted into the database. Following deduplication based on exact address matches, the final consolidated dataset comprised $\totalPropertiesForMLPipeline$ unique property listings, enriched with both API-sourced metadata and PDF-extracted features.

\begin{figure}
\centering
\includegraphics[width=\textwidth]{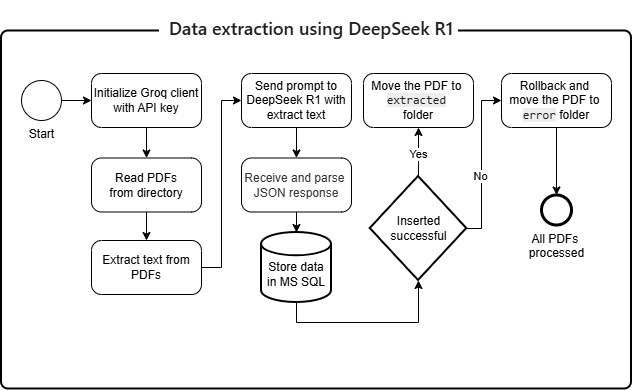}
\caption{Structured data extraction workflow using DeepSeek-R1 LLM}
\label{fig:LLM_based_extraction}
\end{figure}

\subsection{Feature Enrichment and Dataset Consolidation}
The API-sourced property metadata and PDF-extracted features were merged into a unified data frame. Duplicate listings were removed by exact \texttt{full\_address} match. Remaining missing values were imputed uniformly with \texttt{`No'}, treating absence of disclosure as a negative response consistent with the questionnaire format. The \texttt{council\_tax} field was one-hot encoded to convert it into binary indicator columns.

Two composite features were derived. A \texttt{price\_per\_m\textsuperscript{2}} feature was calculated for records with valid area values, with zero-area entries imputed using the overall average price-per-metre ratio. An \texttt{amenity\_score} was constructed as the integer sum of 14 binary property attributes drawn from both API and PDF-extracted sources, producing a value in the range [0, 14]. The numerical features were normalised using \texttt{StandardScaler}, and amenity and preference-weighted features were scaled using \texttt{MinMaxScaler} before model input.

\section{Validation}
\label{sec:valid}
To assess the practical utility of the features extracted by the document processing pipeline described in Section \ref{sec:pipeline}, a series of downstream machine-learning experiments was conducted. These experiments employ a hybrid property recommendation framework as an application testbed. Thus, the recommender system serves as a validation instrument for the extraction pipeline.

Three complementary validation approaches were applied: similarity-based property matching, unsupervised clustering-based segmentation, and Multi-Criteria Decision-Making (MCDM) ranking. Each approaches the feature set from a different analytical angle, covering the range from low-level vector consistency to high-level semantic interpretability.

\subsection{Experimental Setup}
The consolidated dataset of $\totalPropertiesForMLPipeline$ property listings was used as the basis for all validation experiments. Before model training, standard preprocessing was applied. Numerical features (price, area, price\_per\_m\textsuperscript{2}, amenity\_score, bedroom count, bathroom count) were normalised using \texttt{StandardScaler}. Boolean features were already encoded as binary 0/1 values during the consolidation stage, and the \texttt{council\_tax} (i.e., a tax charged in the UK to property occupiers based on the properties' value) categorical feature was transformed via one-hot encoding. The final feature matrix comprised both API-sourced attributes and PDF-extracted features, the latter contributing to the $\pdfAttributesCount$-field structured schema detailed in Section~\ref{sec:pipeline}.

\begin{figure}
\centering
\includegraphics[width=\textwidth]{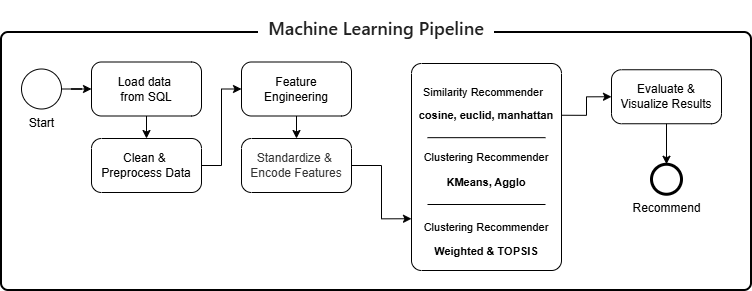}
\caption{Validation pipeline for LLM-based feature extraction}
\label{}
\end{figure}

\subsection{Feature Coherence: Similarity-Based Analysis}
To assess whether the PDF-extracted features produce coherent property representations, cosine similarity was computed pairwise across the full feature matrix, and Top-5 recommendations were generated for a random sample of 50 query properties. Consistency across three similarity metrics, Cosine Similarity, Euclidean Distance, and Manhattan Distance, was evaluated using Jaccard similarity over the Top-5 result sets. Cosine Similarity and Euclidean Distance produced the strongest pairwise agreement ($Jaccard \approx \cosineEuclidean$), indicating that the feature vectors capture genuine relational structure between properties. The lower agreement between Cosine Similarity and Manhattan Distance ($Jaccard \approx \cosineManhattan$) is expected, given the latter's stepwise sensitivity to individual feature magnitudes, and reflects feature diversity rather than noise. These consistency scores would not be achievable if the PDF-extracted features introduced systematic extraction errors or incoherent values. A flat or noisy feature set would produce near-identical or near-random recommendations regardless of the metric applied.

\begin{figure}[H]
\centering
\includegraphics[width=\textwidth]{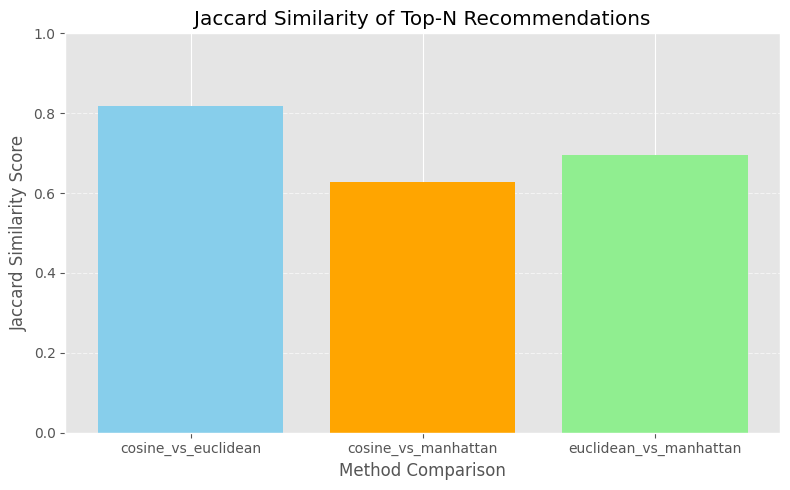}
\caption{Jaccard similarity scores of Top-N recommendations}
\label{Jaccard_Similarity}
\end{figure}

\subsection{Feature Semantic Quality: Cluster Segmentation}
Unsupervised K-Means clustering was applied to the enriched feature matrix to evaluate whether the PDF-extracted features support coherent, interpretable property segmentation. A task that requires semantically meaningful feature values rather than merely structurally valid ones. The optimal cluster count was determined using the Elbow Method on Within-Cluster Sum of Squares (WCSS) and confirmed via Silhouette Score analysis, which peaked at $k = \kMeansClusters$ with a score of $\kMeanSilhouette$.

\begin{table}[H]
\caption{K-Means cluster profiles ($k=2$) — mean feature values per cluster}
\centering
\begin{tabularx}{\textwidth}{llllX}
\toprule
\textbf{Cluster} & \textbf{Price (£)} & \textbf{Area (m$^2$)} & \textbf{Amenity Score} & \textbf{Market Segment} \\
\midrule

\rowcolor{gray!15}
0 & 199,638 & 97.9 & 4.0 & Affordable / smaller properties \\

1 & 229,001 & 115.8 & 3.4 & Premium / larger properties \\

\bottomrule
\end{tabularx}
\label{table:table2}
\end{table}

The resulting segmentation produced two interpretable property-market categories. Cluster 0 captured smaller, more affordably priced properties with moderate amenity profiles. Cluster 1 grouped larger, higher-value properties with higher amenity scores. Table~\ref{table:table2} presents the mean feature values per cluster.

A silhouette score of $\kMeanSilhouette$ peaked at $k=2$. Forcing three or four clusters on this data would produce groupings without meaningful real-world distinction, as the underlying market does not segment cleanly beyond two categories at this scale. The alignment of the two clusters with recognisable market segments, rather than arbitrary statistical groupings, provides evidence that the extracted features carry genuine semantic content. A dataset composed of extraction noise or hallucinated values is unlikely to yield spatially separable, semantically coherent clusters in PCA space.


\begin{figure}[H]
\centering
\includegraphics[width=\textwidth]{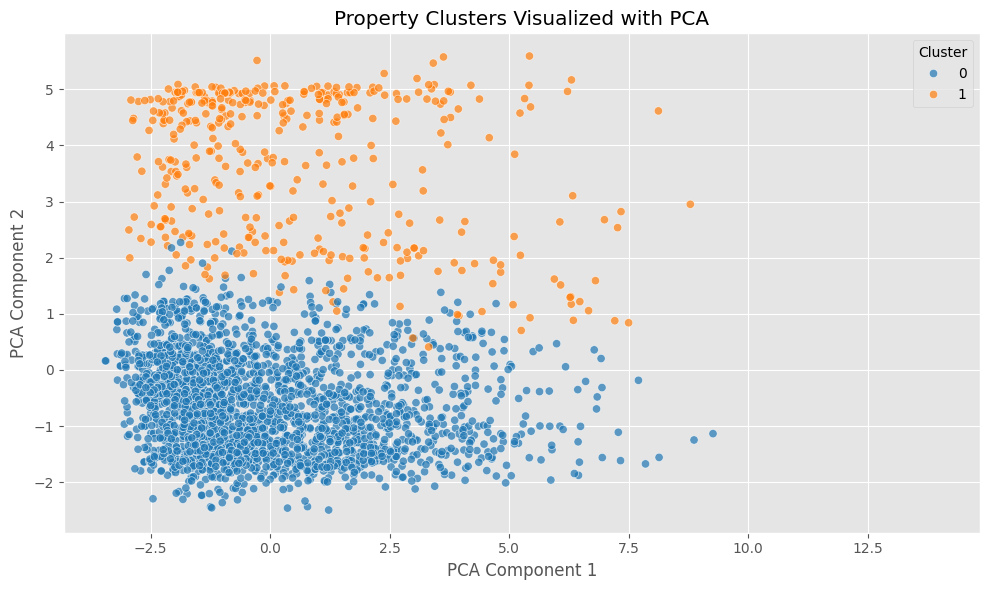}
\caption{PCA visualisation of K-Means clustering (k=2) showing separation between property market segments.}
\label{k_means}
\end{figure}


\subsection{Feature Richness: Multi-Criteria Ranking}
The third validation asks a different question from the first two, not whether the data is consistent or coherent, but whether it is \emph{rich enough} to support differentiated outcomes under distinct ranking objectives. If the extracted features contributed little beyond noise, any two ranking methods applied to the same dataset would produce near-identical results. There would simply not be enough signal for the methods to disagree.

Two MCDM approaches were applied to the same three features (price, floor area, and amenity score) derived from both API and PDF-extracted sources. Weighted Scoring computed a linear combination of MinMax-normalised feature values using user-defined weights, price (30\%), area (40\%), and amenity score (30\%). TOPSIS computed each property's relative closeness to an ideal solution using the formula $s_i = D^-_i / (D^+_i + D^-_i)$, where $D^+$ and $D^-$ are the Euclidean distances to the ideal and anti-ideal solutions, respectively, with price set as a minimisation objective and area and amenity score as maximisation objectives.

The Top-5 lists produced by the two methods shared only 3 properties (Jaccard $\approx$ 0.18). This low overlap is not a weakness. It is evidence that the feature set contains enough variation and structure for two methods with different optimisation logic to reach genuinely different conclusions. A flat or noisy dataset would force both methods to converge on the same handful of extreme outliers. The divergence here shows the data supports nuanced, objective-dependent ranking.

\section{Results}
\label{sec:results}
Table \ref{table:table3} presents the end-to-end performance of the document processing pipeline across all stages. All $\textOnlyPDFs$ documents submitted to the DeepSeek R1 model at the extraction stage produced valid structured JSON output and were successfully inserted into the database. Following deduplication by exact address match, the final data set comprised $\totalPropertiesForMLPipeline$ unique property records enriched with $\pdfAttributesCount$ PDF-extracted features per listing.

\begin{table}[h]
\caption{End-to-end pipeline performance summary}
\centering
\begin{tabularx}{0.98\linewidth}{lXXc}
\toprule
\textbf{Stage} & \textbf{Input} & \textbf{Output} & \textbf{Success Rate} \\
\midrule

\rowcolor{gray!15}
Data acquisition (API) & REST endpoints & $\scrappedProperties$ properties with PDF URLs & 100\% \\

PDF download & URLs & Files & 100\% \\

\rowcolor{gray!15}
PDF classification & $\scrappedProperties$ PDFs & $\textOnlyPDFs$ text\_only & $\textOnlyPDFsPercentage$ eligible \\

LLM extraction (DeepSeek R1) & $\textOnlyPDFs$ text\_only PDFs & $\textOnlyPDFs$ valid JSON records & 100\% \\

\rowcolor{gray!15}
Deduplication & $\textOnlyPDFs$ records & $\totalPropertiesForMLPipeline$ unique listings & - \\

Final dataset & - & $\totalPropertiesForMLPipeline$ records · $\sim$50 features & - \\

\bottomrule
\end{tabularx}
\label{table:table3}
\end{table}

Extraction succeeded on every submitted document, confirming that LLM-based text extraction is viable. The primary bottleneck in the pipeline is the classification stage. $\excudedPDFsPercentage$ of documents (\texttt{scanned} and \texttt{special\_char} classes) could not be processed in the current implementation, representing a meaningful opportunity for pipeline extension through OCR integration and special-character normalisation, as discussed in Section~\ref{sec:concs}.

\subsection{Validation Summary}
Table~\ref{table:table4} consolidates the key quantitative results from all the validation experiments. The results collectively demonstrate that the PDF-extracted feature set supports coherent and consistent ML behaviour across three independent analytical tasks, similarity matching, clustering, and MCDM ranking, providing empirical evidence that the pipeline produces semantically meaningful, ML-ready data.

\begin{table}[h]
\caption{Downstream validation results summary}
\centering
\begin{tabularx}{\textwidth}{XXXX}
\toprule
\textbf{Experiment} & \textbf{Metric} & \textbf{Result} & \textbf{Pipeline Interpretation} \\
\midrule

\rowcolor{gray!15}
Cosine vs Euclidean similarity & Jaccard (Top-5, n=50) & $\cosineEuclidean$ & Strong feature coherence \\

Cosine vs Manhattan similarity & Jaccard (Top-5, n=50) & $\cosineManhattan$ & Expected divergence \\

\rowcolor{gray!15}
K-Means clustering ($k=\kMeansClusters$) & Silhouette Score & $\kMeanSilhouette$ & Coherent segmentation \\

K-Means cluster profiles & Semantic alignment & Affordable vs Premium & Extracted features carry real-world semantic meaning \\

\bottomrule
\end{tabularx}
\label{table:table4}
\end{table}


\subsection{Discussion}
The cosine-Euclidean Jaccard score of $\cosineEuclidean$ shows that the feature vectors are internally consistent. Properties similar under one metric are similar under another, which would not hold if extracted values were noisy or hallucinated. The K-Means silhouette score of $\kMeanSilhouette$ and the cluster profiles, which separate affordable from premium listings along plausible attribute boundaries, suggest the extracted features reflect real distinctions in the property market rather than artefacts of the extraction process.

\subsection{Limitations}
Three limitations of the current study are acknowledged. First, $\excudedPDFsPercentage$ of the PDF documents were excluded from extraction due to \texttt{scanned} or \texttt{special\_char} encoding, introducing potential selection bias in the final dataset. Second, no formal ablation study was conducted comparing downstream model performance with and without PDF-extracted features. Such a comparison would provide the most direct quantitative evidence of the pipeline's additive value over API-only data, and is identified as the primary direction for future work. Third, the validation is entirely quantitative. No user study was conducted to evaluate the perceptual quality of the property representations produced by the enriched dataset.

\section{Conclusion and Future Work}
\label{sec:concs}
\subsection{Conclusions}
This paper presents an end-to-end document intelligence pipeline for the acquisition, classification, and structured extraction of information from real estate listings and questionnaire PDFs.

The pipeline comprises three sequential stages. In stage one, $\scrappedProperties$ PDF documents acquired from the ASPC platform via reverse-engineered REST endpoints were classified into structural categories. Documents classified as \texttt{text\_only} accounted for $\textOnlyPDFsPercentage$ of the corpus and were forwarded to the extraction stage. In the second stage, raw text was extracted from the $\textOnlyPDFs$ eligible text\_only documents using \texttt{pdfplumber}. Lastly, the DeepSeek R1 model, accessed via the Groq inference API, was prompted to extract $\pdfAttributesCount$ predefined property attributes from each document and return them as a structured JSON object. All $\textOnlyPDFs$  submissions were processed successfully, yielding a consolidated dataset of $\totalPropertiesForMLPipeline$ unique property records.

Downstream validation experiments, spanning cosine similarity-based property matching (Jaccard $\approx \cosineEuclidean$), K-Means clustering (silhouette = $\kMeanSilhouette$), and MCDM-based ranking, provide empirical evidence of the quality of pipeline data. The work demonstrates that LLM-based text extraction is feasible and practically viable.

\subsection{Future Work}
Currently, the pipeline excludes \texttt{scanned} and \texttt{special\_char} documents, which together account for approximately $\excudedPDFsPercentage$ of the acquired properties. Extending the pipeline to handle \texttt{scanned} PDFs through integrated OCR preprocessing with noise reduction, deskewing, and adaptive thresholding would reduce the selection bias introduced by restricting extraction to text-only PDFs.

A formal ablation study comparing model performance on API-only features versus the full PDF-enriched feature set would provide the most direct quantitative evidence of the pipeline's additive value, and is identified as the most important immediate extension of this work.

Finally, the pipeline architecture is designed to be domain-agnostic. Future work will evaluate its applicability to other document-rich real estate markets (including commercial and rental property sectors) and to other domains where critical information is similarly trapped in unstructured documents, such as legal case files, medical records, and insurance claim forms.



\subsubsection*{Disclosure of Interests}
The author has no competing interests to declare.

\end{document}